\documentclass{article}

\usepackage[nonatbib, final]{nips_2017}

\usepackage[utf8]{inputenc} 
\usepackage[T1]{fontenc}    
\usepackage{hyperref}       
\usepackage{url}            
\usepackage{booktabs}       
\usepackage{amsfonts}       
\usepackage{nicefrac}       
\usepackage{microtype}      
\usepackage{todonotes}      
\usepackage{algorithm}
\usepackage{algorithmic}  
\usepackage{cite}
\usepackage{amsmath}
\usepackage{bm}
\usepackage{graphicx}
\usepackage{pgfplots}
\usepgfplotslibrary{groupplots}
\usepackage{siunitx}

\title{InterpNET: Neural Introspection for Interpretable Deep Learning}

\author{
  Shane Barratt\thanks{This work was completed while at UC Berkeley.} \\
  Department of Electrical Engineering\\
  Stanford University \\
  Stanford, CA 94305 \\
  \texttt{sbarratt@stanford.edu} \\
}

\begin{document}

\maketitle

\begin{abstract}
Humans are able to explain their reasoning. On the contrary, deep neural networks are not. This paper attempts to bridge this gap by introducing a new way to design interpretable neural networks for classification, inspired by physiological evidence of the human visual system's inner-workings. This paper proposes a neural network design paradigm, termed InterpNET, which can be combined with any existing classification architecture to generate natural language explanations of the classifications. The success of the module relies on the assumption that the network's computation and reasoning is represented in its internal layer activations. While in principle InterpNET could be applied to any existing classification architecture, it is evaluated via an image classification and explanation task. Experiments on a CUB bird classification and explanation dataset show qualitatively and quantitatively that the model is able to generate high-quality explanations. While the current state-of-the-art METEOR score on this dataset is $29.2$, InterpNET achieves a much higher METEOR score of $37.9$. Source code is available online\footnote{https://github.com/sbarratt/interpnet}.
\end{abstract}

\section{Introduction}
An interesting property of deep architectures for supervised learning tasks is that when trained, they are able to extract more and more abstract representations as low-level sensory data flows through computation steps in the network. This property has been verified empirically, and a whole new field called representation learning has been created. A deep classification architecture's success stems from its ability to sequentially extract more abstract and useful features from the previous layer until it extracts the highest-level feature, the class label. Therefore, the intermediate features represent concrete steps in the network's reasoning. If there were a way to extract insight from the activations internal to the network it would be possible to reason about how the network is performing its classifications. It is not plausible for a human to be able to reason about these high-dimensional activations internal to the network, so instead we turn to the idea that another network could operate on these internal activations to describe how the classification network is making its decisions. The idea of having another network generate explanations based on the original classification network's internal activations is InterpNET.

There is inspiration for this idea found in the inner-workings of the human visual system. It turns out that when human subjects imagine visual objects without the actual sensory stimulus (e.g. their eyes are closed), there is still activity in their visual cortex \cite{kastner1999increased}. This means that when we think and reason about images we are actually using the internal representations in our visual cortex. Just as in the brain, InterpNET uses the machinery in its classification network to guide its explanation. Also, research points to evidence that there are feed-forward connections in the visual cortex \cite{lamme1998feedforward}. This means that internal representations in the brain are used further down the pipeline for further reasoning. Similarly, InterpNET essentially uses feed-forward connections from hidden layers in the network to an explanation module down-stream to help reason about the image.

\section{Approach}
\subsection{Problem Statement}

In supervised classification and explanation, one is given supervised trios $(x, y, E(x, y))$, where $x$ is the observation, $y$ is the class and $E$ is a natural-language explanation of the classification based on the observation and resulting class. The goal is to design a model which can accurately assign classes $y$ to observations $x$ along with an explanation $E(x, y)$ of that classification. This problem and resulting approach differs from captioning models in that captioning models are only trained on $(x, E(x))$ pairs and only describe the \emph{observation} and not the network's \emph{reasoning}.

\subsection{InterpNET Architecture}

In principle, the layers of a neural network compute higher and higher-level representations of the parts of the input which are relevant to producing a class label \cite{bengio2013representation}. Therefore, it is reasonable to assume that the relevant aspects to classification are contained in the internal activations of the network. For example, a single ReLU hidden-layer neural network computes the function

$$y = \text{softmax}(W_1 \text{relu} (W_2 x + b_2) + b_1)$$

and has internal activations $f_1, f_2, f_3$:

$$f_1 = x$$
$$f_2 = \text{relu} (W_2 x + b_2)$$
$$f_3 = \text{softmax}(W_1 \text{relu} (W_2 x + b_2) + b_1)$$

Building on this idea, the computation/reasoning of the network can be viewed as the internal activations of the network concatenated into a single feature vector, $r(x) = [f_1, f_2, f_3]$. Then, InterpNET uses this feature vector as input to a language-generating network and trains the language-generator in a supervised fashion to generate explanations $E(x, y)$. The next few sections go through the technical details to make this idea concrete on the problem of fine-grained bird classification.

\subsection{Model Architecture for CUB Dataset}

\paragraph{CUB Dataset}

InterpNET is evaluated on the Caltech-UCSD Birds 200-2011 (CUB) dataset, which has $11,788$ images of birds, each belonging to one of $200$ bird species \cite{WahCUB_200_2011}. Recently, \cite{reed2016learning} collected 10 descriptions for each image which do not describe the content of the image (e.g. ``this is an image of a bird on a tree.") but rather identify class-discriminative visual features (e.g. ``this is a bird with a white belly, brown back and a white eyebrow."). This dataset serves as an important benchmark for models which seek to provide accurate classifications and natural language explanations behind their classifications. InterpNET achieves state-of-the-art on this benchmark dataset. All results presented are on the standard CUB test set.

\begin{figure}[h]
  \centering
  \includegraphics[width=\linewidth]{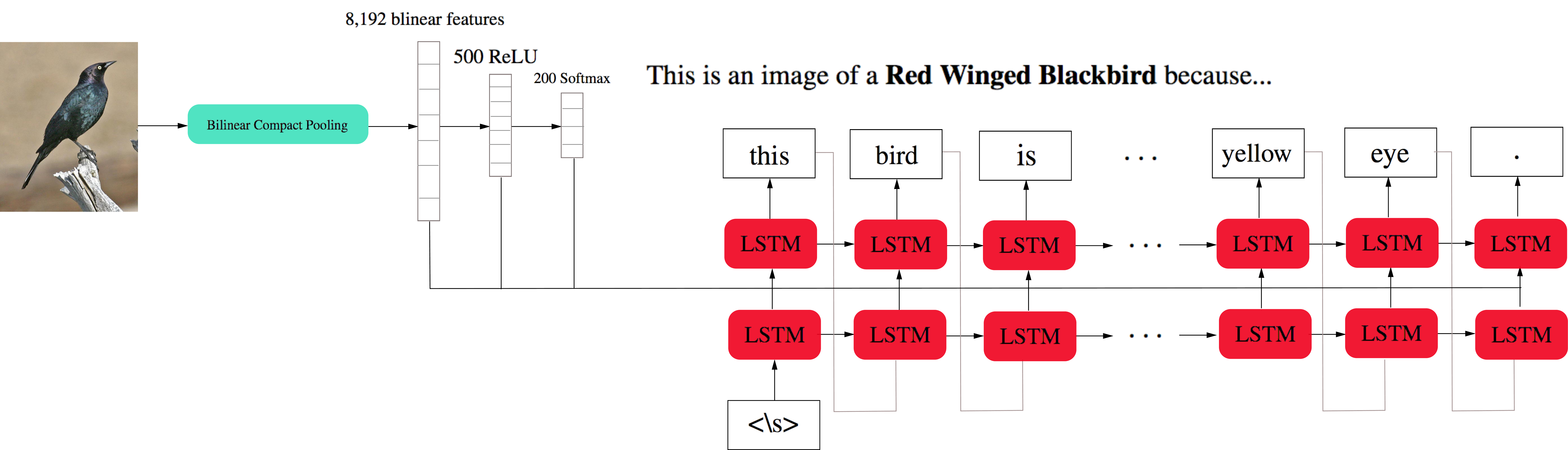}
  \caption{The Model. First, the network extracts 8,192 features using a pre-trained bilinear compact pooling network. Then, it classifies the category of bird using a fully connected network. It then concatenates the internal activations of the fully connected network and provides them as input to a LRCN$_{2f}$ language-generating RNN which is unrolled to produce an explanation of the classification.}
  \label{fig:model}
\end{figure}

Given an observation $x$, the goal is to produce a class $y$ and an explanation $E_t$. In the case of the CUB dataset, the observation $x$ is a $M$x$N$x$3$ RGB image and the explanation $E_t$ is a vector of word indexes, ending with a terminal word index (a period). A dictionary $D$ maps word indexes to English words, and includes a start word and terminal word. The variable $\theta$ represents the model parameters. The classifier distribution is represented as $\Pr(y|x; \theta)$ and the explanation distribution is represented as $\Pr(E_t|E_{t-1}, r(x); \theta)$. The full model is summarized in Figure \ref{fig:model}.

For the CUB dataset, each image is preprocessed into a 8,192 dimensional feature, the second to last layer of the compact bilinear pooling network \cite{gao2016compact} which was pre-trained on the CUB dataset. These features then get fed into a series of hidden ReLU layers (for illustration, 1 is shown in the Figure) then a classification softmax layer to model $\Pr(y|x; \theta)$. Let the concatenation of the resulting feature layers of the classification network be $r(x)$.

A language-generating Long-Short Term Memory Network (LSTM) is used to represent $\Pr(E_t|E_{t-1}, r(x); \theta)$. More specifically, InterpNET uses a two-layer LSTM where $r(x)$ is concatenated to the input to the second LSTM. InterpNET's language generator is equivalent to LRCN$_{2f}$ which achieved the highest caption-to-image retrieval performance in work surveying different recurrent architectures for captioning \cite{donahue2015long}.

The loss function for the Classifier, $L_C$, is the cross-entropy loss between the output class probabilities and the actual class probabilities. The loss function for the explanation module, $L_E$, is the cross-entropy loss between the output sentence probabilities and the desired sentence.

\subsection{Training Procedure}

Because there are two separate but connected neural networks in the model that need to be trained, there are many possible variants to the overall training procedure. When gradient descent is run on the explanation module, the parameters of the classification model affect the explanation module and thus the gradient $\nabla_{\theta} L_E$ includes terms from the classification model. Therefore, in this paper, the gradient is stopped at $r(x)$ to avoid modifying the classifier parameters and thus sacrificing accuracy. The final training routine involves training the classifier to convergence and then the explainer to convergence, and is summarized in Algorithm \ref{alg2} in the Appendix. Both networks are trained using stochastic gradient descent (SGD) with momentum, specifically the ADAM algorithm \cite{kingma2014adam}. Alternated training procedures were investigated, but this one was the simplest and worked the best.

\section{Experiments}
\subsection{Evaluation Metrics}

InterpNET's explanations were evaluated using a variety of automated metrics. The metrics include the bilingual evaluation understudy (BLEU) score from machine translation, the Automatic NT Translation Metric (METEOR) and Consensus-based Image Description Evaluation (CIDEr). BLEU measures the similarity of sentences based on an averaged percentage of n-gram matches \cite{bleu} and is one of the first metrics to highly correlate with human judgments of similarity \cite{evaluating_bleu}. METEOR does a similar evaluation as BLEU, but uses pre-trained word embeddings to semantically evaluate the similarity between words \cite{denkowski:lavie:meteor-wmt:2014}. CIDEr measures similarity between generated sentences to reference explanations by counting TF-IDF weighted n-grams \cite{vedantam2015cider}. CIDEr rewards uncommon sentences which are used correctly.

\subsection{Experiments Setup}

Multiple architectures were tested and Table~\ref{tab:results} shows the automated metrics and classification accuracy on the standard CUB test set. For all metrics, higher is better. The five architectures evaluated are: (1) InterpNET$_0$: only the class probabilites generated by the classification network are fed to the language-generating RNN, (2) InterpNET$_1$: the classification network has one hidden layer and all layer activations are fed to the language-generating RNN, (3) InterpNET$_2$: the classification network has two hidden layers and all layer activations are fed to the language-generating RNN, (4) InterpNET$_3$: the classification network has three hidden layers and all layer activations are fed to the language-generating RNN, (5) Captioning: only the 8,192 dimensional image feature is fed to the language-generating RNN and (6) Generating Visual Explanations (the baseline).

\subsection{Quantitative Results}

\begin{table}[t]
  \caption{Results. Explanation metrics and Classification Accuracy for a variety of models. InterpNET$_2$ achieves the highest metrics, except for classification accuracy. Higher is better for all metrics.}
  \label{tab:results}
  \centering
  \begin{tabular}{lllll}
    \toprule
     & METEOR & BLEU & CIDer & Classification Accuracy \\
    \midrule
    InterpNET$_0$ (output only) & $35.0$ & $55.6$ & $68.3$ & $81.3\%$ \\
    InterpNET$_1$ (1 hidden layer) & $36.1$ & $58.7$ & $73.5$ & $\textbf{81.5\%}$ \\
    InterpNET$_2$ (2 hidden layers) & $\textbf{37.9}$ & $\textbf{62.3}$ & $\textbf{82.1}$ & $79\%$ \\
    InterpNET$_3$ (3 hidden layers) & $31.7$ & $47.3$ & $54.3$ & $76.7\%$ \\
    Captioning (input only) & $32.2$ & $48.2$ & $55.5$ & $81.3\%$ \\
    Generating Visual Explanations (baseline) & $29.2$ & n/a & $56.7$ & n/a \\
    \bottomrule
  \end{tabular}
\end{table}

Table~\ref{tab:results} shows the quantitative experiment results. All approaches evaluated in this paper have higher METEOR and CIDer scores than the state-of-the-art baseline model \cite{hendricks2016generating}. Thus, InterpNET is now the state-of-the-art for generating visual explanations.

The highest performing network across all metrics was InterpNET$_2$, which employed two hidden layers. There was also a trade-off between the number of hidden layers in the classification network and the explanation metrics; too few lead to not enough information and too many lead to over-fitting. More hidden layers also led to a lower classification accuracy, likely because of high model expressivity and thus over-fitting.

All InterpNET instantiations, except for InterpNET$_3$, had higher metrics than the captioning architecture which means that InterpNET's architecture is superior for the task of explaining a network's classifications. It also provides substantial evidence backing the claim that a representation of the reasoning behind the network's classification is contained in its internal activations.

Surprisingly, InterpNET$_0$, which acts only on the class probabilities, outperforms captioning and is almost at the level of the other networks. This means that the statistics of the class probabilites outputted by the network are well correlated with the explanations, as one would expect. However, it achieves a low CIDer score of $68.3$, likely because the network memorizes the best explanation for each class making its sentences unoriginal.

\section{Conclusion}

This paper introduces a general neural network module which can be combined with any existing classification architecture to generate natural language explanations of the network's classifications provided one has supervised explanation data. InterpNET's classifications are highly accurate and interpretable at the same time as demonstrated by quantitative and qualitative analysis of experiments on a bird classification+explanation dataset. InterpNET achieves a METEOR score of $37.9$ on the CUB test set, making it state-of-the-art in the visual explanation task. The model is able to use the information extracted from a trained classifier to produce excellent explanations and is a sizable step towards interpretable deep neural network models.

Future work involves testing the InterpNET module on different classification architectures and on domains outside of computer vision (for example in skin cancer classification and fraud detection). Further extensions also include more complex language-generating architectures with attention and adversarial-based training architectures. Making complex neural networks interpretable by humans is one of the main doubts practitioners have, and thus is an important problem to address moving forward.

\small

\bibliography{mybib}{}

\begin{thebibliography}{10}

\bibitem{kastner1999increased}
Sabine Kastner, Mark~A Pinsk, Peter De~Weerd, Robert Desimone, and Leslie~G
  Ungerleider.
\newblock Increased activity in human visual cortex during directed attention
  in the absence of visual stimulation.
\newblock {\em Neuron}, 22(4):751--761, 1999.

\bibitem{lamme1998feedforward}
Victor~AF Lamme, Hans Super, and Henk Spekreijse.
\newblock Feedforward, horizontal, and feedback processing in the visual
  cortex.
\newblock {\em Current opinion in neurobiology}, 8(4):529--535, 1998.

\bibitem{bengio2013representation}
Yoshua Bengio, Aaron Courville, and Pascal Vincent.
\newblock Representation learning: A review and new perspectives.
\newblock {\em IEEE transactions on pattern analysis and machine intelligence},
  35(8):1798--1828, 2013.

\bibitem{WahCUB_200_2011}
C.~Wah, S.~Branson, P.~Welinder, P.~Perona, and S.~Belongie.
\newblock {The Caltech-UCSD Birds-200-2011 Dataset}.
\newblock Technical Report CNS-TR-2011-001, California Institute of Technology,
  2011.

\bibitem{reed2016learning}
Scott Reed, Zeynep Akata, Honglak Lee, and Bernt Schiele.
\newblock Learning deep representations of fine-grained visual descriptions.
\newblock In {\em Proceedings of the IEEE Conference on Computer Vision and
  Pattern Recognition}, pages 49--58, 2016.

\bibitem{gao2016compact}
Yang Gao, Oscar Beijbom, Ning Zhang, and Trevor Darrell.
\newblock Compact bilinear pooling.
\newblock In {\em Proceedings of the IEEE Conference on Computer Vision and
  Pattern Recognition}, pages 317--326, 2016.

\bibitem{donahue2015long}
Jeffrey Donahue, Lisa Anne~Hendricks, Sergio Guadarrama, Marcus Rohrbach,
  Subhashini Venugopalan, Kate Saenko, and Trevor Darrell.
\newblock Long-term recurrent convolutional networks for visual recognition and
  description.
\newblock In {\em Proceedings of the IEEE conference on computer vision and
  pattern recognition}, pages 2625--2634, 2015.

\bibitem{kingma2014adam}
Diederik Kingma and Jimmy Ba.
\newblock Adam: A method for stochastic optimization.
\newblock {\em arXiv preprint arXiv:1412.6980}, 2014.

\bibitem{bleu}
Kishore Papineni, Salim Roukos, Todd Ward, and Wei-Jing Zhu.
\newblock Bleu: a method for automatic evaluation of machine translation.
\newblock In {\em Proceedings of the 40th annual meeting on association for
  computational linguistics}, pages 311--318. Association for Computational
  Linguistics, 2002.

\bibitem{evaluating_bleu}
Chris Callison-Burch, Miles Osborne, and Philipp Koehn.
\newblock Re-evaluation the role of bleu in machine translation research.
\newblock In {\em EACL}, volume~6, pages 249--256, 2006.

\bibitem{denkowski:lavie:meteor-wmt:2014}
Michael Denkowski and Alon Lavie.
\newblock Meteor universal: Language specific translation evaluation for any
  target language.
\newblock In {\em Proceedings of the EACL 2014 Workshop on Statistical Machine
  Translation}, 2014.

\bibitem{vedantam2015cider}
Ramakrishna Vedantam, C~Lawrence~Zitnick, and Devi Parikh.
\newblock Cider: Consensus-based image description evaluation.
\newblock In {\em Proceedings of the IEEE Conference on Computer Vision and
  Pattern Recognition}, pages 4566--4575, 2015.

\bibitem{hendricks2016generating}
Lisa~Anne Hendricks, Zeynep Akata, Marcus Rohrbach, Jeff Donahue, Bernt
  Schiele, and Trevor Darrell.
\newblock Generating visual explanations.
\newblock In {\em European Conference on Computer Vision}, pages 3--19.
  Springer International Publishing, 2016.

\bibitem{russakovsky2015imagenet}
Olga Russakovsky, Jia Deng, Hao Su, Jonathan Krause, Sanjeev Satheesh, Sean Ma,
  Zhiheng Huang, Andrej Karpathy, Aditya Khosla, Michael Bernstein, et~al.
\newblock Imagenet large scale visual recognition challenge.
\newblock {\em International Journal of Computer Vision}, 115(3):211--252,
  2015.

\bibitem{rowley1998neural}
Henry~A Rowley, Shumeet Baluja, and Takeo Kanade.
\newblock Neural network-based face detection.
\newblock {\em IEEE Transactions on pattern analysis and machine intelligence},
  20(1):23--38, 1998.

\bibitem{bahdanau2014neural}
Dzmitry Bahdanau, Kyunghyun Cho, and Yoshua Bengio.
\newblock Neural machine translation by jointly learning to align and
  translate.
\newblock {\em arXiv preprint arXiv:1409.0473}, 2014.

\bibitem{rush2015neural}
Alexander~M Rush, Sumit Chopra, and Jason Weston.
\newblock A neural attention model for abstractive sentence summarization.
\newblock {\em arXiv preprint arXiv:1509.00685}, 2015.

\bibitem{showandtell2014}
Oriol Vinyals, Alexander Toshev, Samy Bengio, and Dumitru Erhan.
\newblock Show and tell: A neural image caption generator.
\newblock In {\em Proceedings of the IEEE Conference on Computer Vision and
  Pattern Recognition}, pages 3156--3164, 2015.

\bibitem{antol2015vqa}
Stanislaw Antol, Aishwarya Agrawal, Jiasen Lu, Margaret Mitchell, Dhruv Batra,
  C~Lawrence~Zitnick, and Devi Parikh.
\newblock Vqa: Visual question answering.
\newblock In {\em Proceedings of the IEEE International Conference on Computer
  Vision}, pages 2425--2433, 2015.

\end{thebibliography}
\bibliographystyle{unsrt}

\section{Appendix}

\subsection{Related Work}
Many recent advances in machine learning have come from deep learning, which employs a model composed of multiple non-linear transformations and gradient-based training to fit the underlying parameters. For vision tasks, deep convolutional networks have achieved state-of-the-art in object detection \cite{russakovsky2015imagenet}, face detection \cite{rowley1998neural} and many others. For language understanding tasks, deep networks have also achieved state-of-the-art in machine translation \cite{bahdanau2014neural}, summarization \cite{rush2015neural} and many others. At the intersection of vision and language there have been breakthrough results in captioning \cite{showandtell2014}, visual question answering \cite{antol2015vqa} and many others. 

The most closely related work to this is on generating visual explanations \cite{hendricks2016generating}. The authors propose a method for deep visual explanations which uses a standard captioning model but also incorporates a loss function which rewards class specificity. The experimental validation of InterpNET is largely based on the machinery they used for fine-grained bird classification. InterpNET, which is much simpler, in fact outperforms the method in \cite{hendricks2016generating} on measures of both accuracy and class-specificity.

\subsection{Qualitative Results}

\begin{figure}
  \centering
  \includegraphics[width=\linewidth]{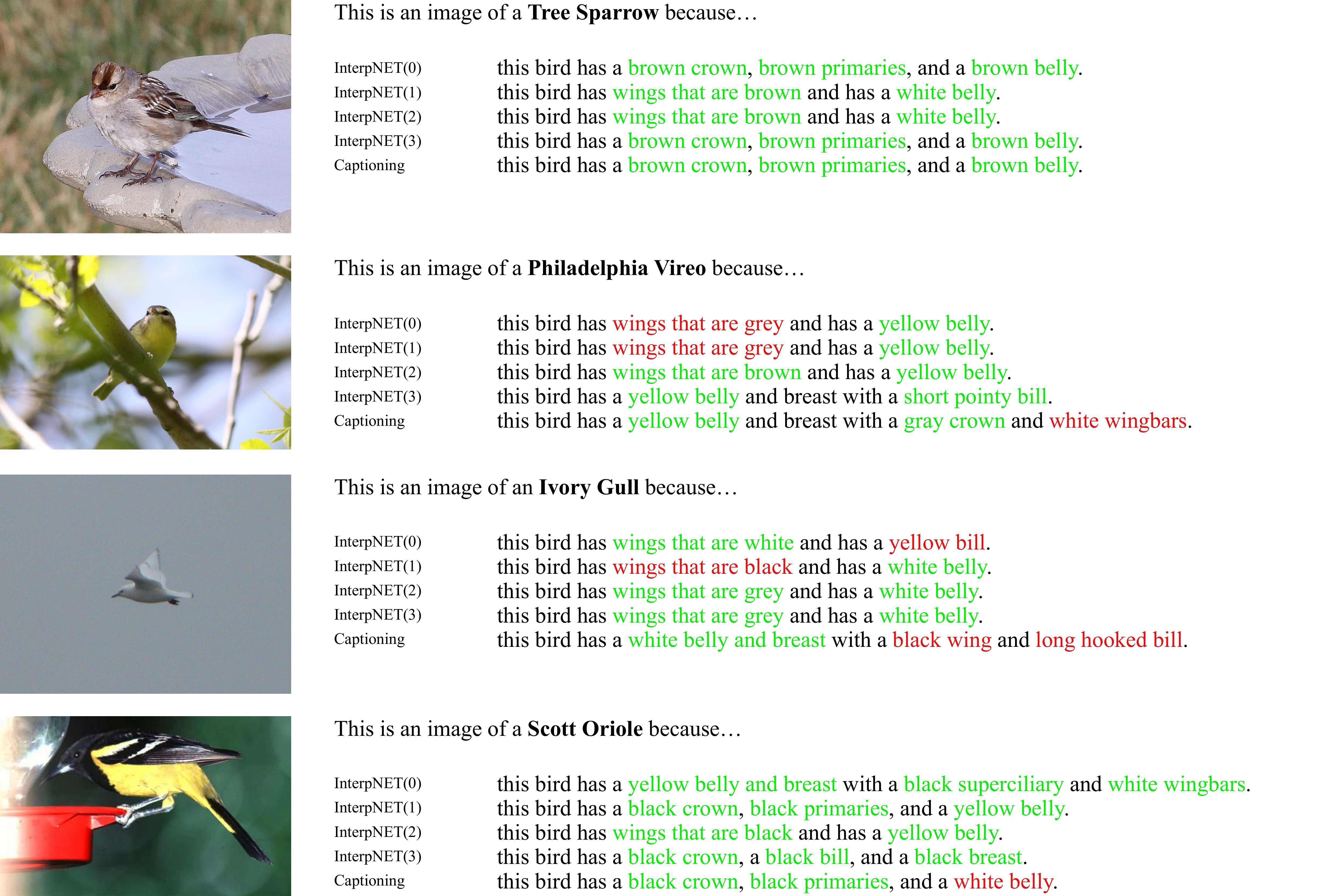}
  \caption{Example classifications and explanations. Green and red text signify a valid and invalid descriptor respectively.}
  \label{fig:examples}
\end{figure}

Figure \ref{fig:examples} shows example explanations for images in the CUB test set for different architectures. The explanations accurately identify discriminating features in the image and provide reasoning behind the network's classification. The descriptors are green or red based on the image they are describing; green text signifies accurate descriptors and red text signifies innacurate descriptors. All of the models' explanations match the image well, but the InterpNET models seem to be the most accurate. The captioning descriptions provide more descriptors but seem to be innacurate a lot of the time, most likely because the captioning model is only looking at the image and does not have class-specific knowledge like the others.

\subsection{Training Procedure}

\begin{algorithm}
  \caption{InterpNET Training Procedure.}
  \label{alg2}
  \begin{algorithmic}
    \WHILE{epochs $<$ some number}
      \STATE Update Classifier parameters $\theta_C$ using ADAM on $L_C$ with early stopping
    \ENDWHILE
    \WHILE{epochs $<$ some number}
      \STATE Update Explainer parameters $\theta_D$ using ADAM on $L_E$ with early stopping
    \ENDWHILE
  \end{algorithmic}
\end{algorithm}

\end{document}